\documentclass[11pt]{article}
\usepackage{acl}
\usepackage{float}

\usepackage{latexsym}

\usepackage{microtype}
\usepackage{inconsolata}
\usepackage{graphicx}
\usepackage{booktabs}
\usepackage{amsmath}
\usepackage{multirow}
\usepackage{tikz}
\usepackage{tcolorbox}
\tcbuselibrary{skins}

\usepackage{fontspec}
\usepackage{polyglossia}
\usepackage{enumitem}
\setmainlanguage{english}
\setotherlanguage{persian}
\newfontfamily\persianfont[Script=Arabic,Scale=1.0]{FreeSerif.ttf}

\newcommand{\pcomma}{{\persianfont ،}}
\newcommand{\pquestion}{{\persianfont ؟}}
\newcommand{\psemicolon}{{\persianfont ؛}}

\newtcolorbox{persianbox}[1][]{
  colback=gray!5,
  colframe=gray!40,
  boxrule=0.5pt,
  rounded corners,
  fontupper=\persianfont,
  #1
}

\title{PersianPunc: A Large-Scale Dataset and BERT-Based Approach for Persian Punctuation Restoration}

\author{
Mohammad Javad Ranjbar Kalahroodi\textsuperscript{1}, 
Heshaam Faili\textsuperscript{1}, 
Azadeh Shakery\textsuperscript{1,2} \\[6pt]
\textsuperscript{1}University of Tehran, Tehran, Iran \\
\textsuperscript{2} Institute for Research in Fundamental Sciences (IPM), Tehran, Iran \\
\texttt{\{mohammadjranjbar, hfaili, shakery\}@ut.ac.ir}
}

\begin{document}

\maketitle
\begin{abstract}
Punctuation restoration is essential for improving the readability and downstream utility of automatic speech recognition (ASR) outputs, yet remains underexplored for Persian despite its importance. We introduce \textbf{PersianPunc}, a large-scale, high-quality dataset of 17 million samples for Persian punctuation restoration, constructed through systematic aggregation and filtering of existing textual resources. We formulate punctuation restoration as a token-level sequence labeling task and fine-tune ParsBERT to achieve strong performance. Through comparative evaluation, we demonstrate that while large language models can perform punctuation restoration, they suffer from critical limitations: over-correction tendencies that introduce undesired edits beyond punctuation insertion (particularly problematic for speech-to-text pipelines) and substantially higher computational requirements. Our lightweight BERT-based approach achieves a macro-averaged F1 score of 91.33\% on our test set while maintaining efficiency suitable for real-time applications. We make our \href{https://huggingface.co/datasets/MohammadJRanjbar/persian-punctuation-restoration}{dataset} and \href{https://huggingface.co/MohammadJRanjbar/parsbert-persian-punctuation}{model} publicly available to facilitate future research in Persian NLP and provide a scalable framework applicable to other morphologically rich, low-resource languages.\footnotemark

\end{abstract}

\footnotetext{Our resources are publicly available: \href{https://huggingface.co/datasets/MohammadJRanjbar/PersianPunc}{full dataset (17M samples)}, \href{https://huggingface.co/datasets/MohammadJRanjbar/persian-punctuation-restoration}{training subset}, and \href{https://mohammadjranjbar.github.io/PersianPunc/}{fine-tuned model}.}
\section{Introduction}
Punctuation restoration represents an essential task in natural language processing, particularly for languages with limited computational resources. The absence of punctuation in raw text---whether from automatic speech recognition, informal digital communication, or historical documents---severely impacts the performance of downstream NLP tasks including machine translation, text summarization, and sentiment analysis.

\definecolor{academicnavy}{HTML}{2c3e50}
\definecolor{slategray}{HTML}{34495e}
\definecolor{steelblue}{HTML}{7f8c8d}
\definecolor{researchblue}{HTML}{3498db}
\definecolor{deepresearch}{HTML}{2980b9}
\definecolor{academicpurple}{HTML}{8e44ad}
\definecolor{successgreen}{HTML}{27ae60}
\definecolor{warningorange}{HTML}{f39c12}
\definecolor{errorred}{HTML}{e74c3c}
\definecolor{cloudwhite}{HTML}{ecf0f1}
\definecolor{silvergray}{HTML}{bdc3c7}
\definecolor{concretegray}{HTML}{95a5a6}

\begin{figure}[t]
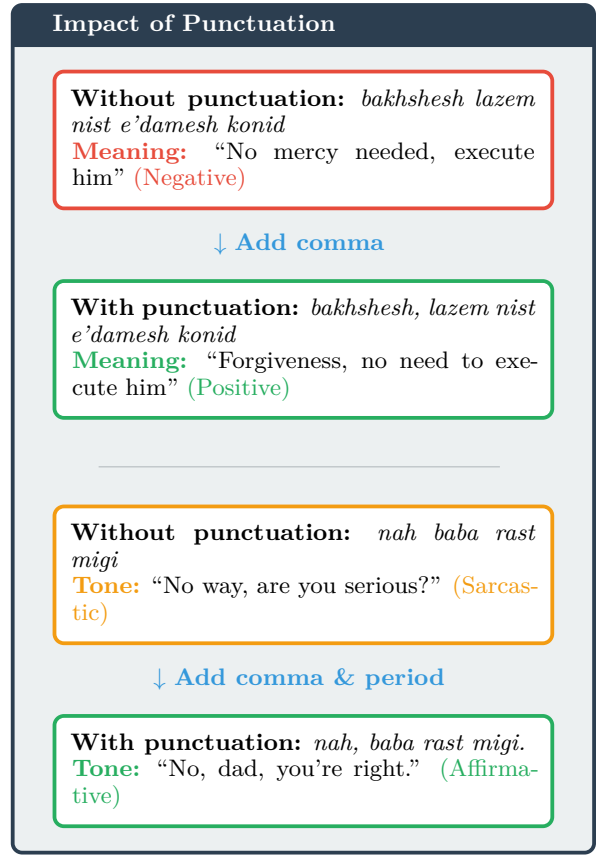

\centering
\small
\begin{tcolorbox}[colback=cloudwhite, colframe=academicnavy, width=\columnwidth,
                  title=\textbf{Impact of Punctuation},
                  fonttitle=\footnotesize\bfseries]
\begin{tcolorbox}[colback=white, colframe=errorred, width=\linewidth,
                  left=3pt, right=3pt, top=3pt, bottom=3pt]
\textbf{Without punctuation:} \textit{bakhshesh lazem nist e'damesh konid}\\
\textcolor{errorred}{\textbf{Meaning:}} ``No mercy needed, execute him'' \textcolor{errorred}{(Negative)}
\end{tcolorbox}
\vspace{3pt}
\centering{\textcolor{researchblue}{$\downarrow$ \textbf{Add comma}}}
\vspace{3pt}
\begin{tcolorbox}[colback=white, colframe=successgreen, width=\linewidth,
                  left=3pt, right=3pt, top=3pt, bottom=3pt]
\textbf{With punctuation:} \textit{bakhshesh, lazem nist e'damesh konid}\\
\textcolor{successgreen}{\textbf{Meaning:}} ``Forgiveness, no need to execute him'' \textcolor{successgreen}{(Positive)}
\end{tcolorbox}
\vspace{6pt}
\centering{\textcolor{silvergray}{\rule{0.8\linewidth}{0.5pt}}}
\vspace{6pt}
\begin{tcolorbox}[colback=white, colframe=warningorange, width=\linewidth,
                  left=3pt, right=3pt, top=3pt, bottom=3pt]
\textbf{Without punctuation:} \textit{nah baba rast migi}\\
\textcolor{warningorange}{\textbf{Tone:}} ``No way, are you serious?'' \textcolor{warningorange}{(Sarcastic)}
\end{tcolorbox}
\vspace{3pt}
\centering{\textcolor{researchblue}{$\downarrow$ \textbf{Add comma \& period}}}
\vspace{3pt}
\begin{tcolorbox}[colback=white, colframe=successgreen, width=\linewidth,
                  left=3pt, right=3pt, top=3pt, bottom=3pt]
\textbf{With punctuation:} \textit{nah, baba rast migi.}\\
\textcolor{successgreen}{\textbf{Tone:}} ``No, dad, you're right.'' \textcolor{successgreen}{(Affirmative)}
\end{tcolorbox}
\end{tcolorbox}
\caption{Persian punctuation restoration dramatically affects semantic interpretation. Minimal punctuation changes transform sentence meaning from negative to positive sentiment.}
\label{fig:punctuation_impact}
\end{figure}

Despite the growing maturity of Persian NLP, punctuation restoration has received limited attention compared to other languages. The critical importance of punctuation in Persian is evidenced by dramatic semantic changes that occur with minimal punctuation modifications, as shown in Figure~\ref{fig:punctuation_impact}. Existing Persian studies have been constrained by small-scale datasets, domain-specific applications, or lack of publicly available models, highlighting the critical need for comprehensive approaches that can handle the full complexity of Persian text across diverse domains and writing styles.

This work addresses these challenges through a comprehensive approach to Persian punctuation restoration using fine-tuned BERT models and large-scale dataset curation. We present a dataset curation methodology that systematically aggregates multiple Persian text sources, resulting in a high-quality corpus for training robust punctuation restoration models.
Our main contributions are:
\begin{itemize}
\item We present \textbf{PersianPunc}, a large-scale Persian punctuation restoration dataset containing 17 million filtered and deduplicated samples spanning diverse domains, sourced from six complementary corpora covering both formal and informal Persian text.
\item We provide a systematic dataset curation framework including detailed preprocessing, quality filtering, and train/validation/test splits, with comprehensive analysis of punctuation distribution patterns in Persian.
\item We achieve strong performance on Persian punctuation restoration with a fine-tuned ParsBERT model, demonstrating competitive results compared to large language models while requiring significantly lower computational resources and avoiding over-correction issues.
\end{itemize}

\section{Related Work}
\label{sec:related_work}
Our research is situated at the intersection of punctuation restoration, Transformer-based NLP, and Persian text processing. This section reviews the evolution of methodologies for this task, starting from general approaches and progressively narrowing the focus to the specific challenges and prior work in the Persian language.

\subsection{Punctuation Restoration as a Sequence Modeling Task}
Historically, punctuation restoration was tackled with statistical methods, including n-gram language models \cite{beeferman1998cyberpunc, gravano2009restoring} and models incorporating prosodic features from speech to predict boundaries \cite{christensen2001punctuation, kim2003combined}. These early systems laid the groundwork but were often limited by the scope of their handcrafted features and statistical models.

The advent of deep learning marked a significant shift. Recurrent Neural Networks (RNNs), particularly models using Bidirectional Long Short-Term Memory (BiLSTM) units, became the standard, framing the problem as a sequence labeling task where each token is classified with a punctuation mark (or none) \cite{xu2016investigating}. This paradigm was often enhanced with Convolutional Neural Networks (CNNs) to capture local character-level features \cite{tundik2018joint, zelasko2018punctuation}, leading to substantial performance gains over classical methods. Our work follows this successful sequence labeling formulation, leveraging a more powerful neural architecture.

\subsection{Transformer-Based Approaches for Punctuation Restoration}
The introduction of the Transformer architecture \cite{vaswani2017attention} and pre-trained language models like BERT \cite{devlin2019bert} revolutionized NLP. For punctuation restoration, fine-tuning BERT-based models quickly became the state-of-the-art approach in high-resource languages, demonstrating superior performance in capturing long-range dependencies crucial for understanding sentence structure and punctuation placement \cite{courtland2020efficient, yi2020focal, nagy2021automatic}. These models are typically lightweight and efficient, making them suitable for real-time applications like ASR post-processing.

More recently, Large Language Models (LLMs) have demonstrated impressive capabilities in a zero-shot or few-shot capacity for various text generation and correction tasks \cite{brown2020language}. However, their application to focused tasks like punctuation restoration comes with potential drawbacks, including high computational inference costs and a tendency for over-correction, where they may alter the source text beyond simply adding punctuation. A key part of our contribution is to rigorously evaluate these trade-offs against a fine-tuned, specialized model.

\subsection{State of Persian Text Processing and Punctuation}
\label{sec:persian_related}
Early work on Persian punctuation restoration was pioneering but limited in scale. \citet{hosseini2017creating} introduced the first known corpus for this task, achieving an F1-score of 69.0\% with a Conditional Random Field (CRF) model. While foundational, this work highlighted the need for larger and more diverse datasets and more powerful models.

More recently, \citet{farokhshad2021virapart} proposed ViraPart, a multi-task text refinement framework for Persian that handles punctuation restoration, Zero-Width Non-Joiner (ZWNJ) recognition, and Ezafe construction. Using ParsBERT \cite{farahani2020parsbert}, they achieved a strong F1-score of 92.13\% for punctuation on the Bijankhan corpus \cite{bijankhan2004role}. Their work demonstrated the effectiveness of Transformer-based models for Persian text refinement. However, ViraPart focuses on multiple text refinement tasks simultaneously, and was evaluated on a smaller, single-domain corpus.

Despite these advances, critical gaps remain. First, there is a lack of a large-scale, publicly available dataset specifically curated for punctuation restoration across diverse domains. Most existing efforts rely on smaller or general-purpose corpora like Bijankhan. Second, the capabilities and limitations of modern LLMs for Persian punctuation restoration have not been systematically studied, particularly regarding over-correction behavior.

\section{Methodology}

\subsection{Dataset Construction}

\subsubsection{Data Sources and Collection Strategy}
\label{sec:data_sources}
We construct a comprehensive Persian punctuation restoration dataset by systematically aggregating high-quality corpora spanning diverse domains and registers. Our multi-source approach addresses the linguistic diversity challenges of Persian NLP through careful curation. We selected source datasets through manual inspection, verifying that at least 100 random samples from each contained proper punctuation usage.

Our dataset combines sources across two primary categories:

\textbf{Formal Academic Text:} Bijankhan-Peykare Corpus \cite{bijankhan2011bijankhan}, Persian Medical QA \cite{ranjbar2023persian}, and Persian Wikipedia \cite{maral2023persian} provide standardized punctuation patterns in formal contexts, covering literary, medical, and encyclopedic domains.

\textbf{Contemporary Informal Text:} Persian Telegram Channels \cite{shojaei2023persian}, Farsi Stories \cite{farsitiny2023stories}, and Blog Dataset V2 \cite{rohanblog2023persian} capture modern conversational patterns and varied punctuation usage, representing social media, narrative fiction, and personal blogging styles.

\subsubsection{Preprocessing and Quality Control}
\label{sec:preprocessing}

\paragraph{Normalization Pipeline}
All texts undergo systematic preprocessing to ensure consistency:
\begin{enumerate}
    \item \textbf{Punctuation standardization:} English punctuation marks (comma, semicolon, question mark) are converted to their Persian equivalents (Persian comma \pcomma, Persian semicolon \psemicolon, Persian question mark \pquestion).
    \item \textbf{Character filtering:} Non-Persian characters are removed while preserving common Persian script variants and Arabic letters used in Persian text.
    \item \textbf{Whitespace normalization:} Multiple spaces are collapsed, and leading/trailing whitespace is removed.
\end{enumerate}

\paragraph{Sentence Segmentation and Filtering}
We first segment each source corpus into sentence-level units using end-of-sentence punctuation marks (period, exclamation mark, question mark). Each candidate sentence then undergoes multi-stage filtering:

\begin{itemize}
    \item \textbf{Structural requirements:} Minimum length of 10 characters; at least two target punctuation marks from the set \{\texttt{.}, \texttt{\pcomma}, \texttt{؟}, \texttt{!}, \texttt{;}, \texttt{:}\}; proper sentence termination with period, exclamation mark, or question mark.
    \item \textbf{Content filtering:} Removal of sentences containing URLs, email addresses, social media handles (@ mentions), emojis, excessive special symbols (more than 20\% non-alphabetic characters), or substantial mixed-language content (more than 30\% non-Persian text).
    \item \textbf{Linguistic quality:} Pattern-based detection and removal of repetitive punctuation (e.g., ``....'', ``!!!!''), enumerative sequences (numbered lists, bullet points), and fragmented text (sentences with more than 50\% single-character tokens).
\end{itemize}

\paragraph{Rationale for Filtering Criteria}
The requirement for at least two punctuation marks ensures that samples present meaningful punctuation restoration challenges beyond simple sentence termination. While this filtering criterion does exclude simple sentences (which are underrepresented in formal Persian writing), it ensures the dataset focuses on the core challenge of internal sentence punctuation, which is critical for ASR applications where sentence boundaries are often detected separately. We acknowledge this as a dataset characteristic rather than a limitation, as it creates a focused benchmark for comma, colon, and question mark insertion---the most challenging and impactful aspects of Persian punctuation restoration. Future work could address simple sentence coverage through stratified sampling or separate evaluation sets.

\subsubsection{Deduplication and Dataset Splitting}
\label{sec:deduplication}
To ensure dataset quality and prevent data leakage, we perform exact deduplication across all source corpora. Due to the large dataset size (initial pool of over 20 million samples), we implement an efficient SHA-256 hash-based deduplication strategy with whitespace normalization. Each sentence is normalized (lowercased, whitespace-collapsed) before hashing to detect duplicates that differ only in formatting.

After deduplication, our final dataset contains 17,102,014 unique samples. For model training and evaluation, we randomly sample a 1M subset stratified by source corpus. This subset is split into training (989,000 samples), validation (10,000 samples), and test (1,000 samples) sets. The sampling strategy maintains the source distribution proportions to ensure representativeness across domains.

\subsection{Dataset Statistics and Punctuation Analysis}
We conducted a comprehensive punctuation analysis on the complete dataset of 17,102,014 samples to understand the characteristics of Persian punctuation usage in our corpus.

\subsubsection{Punctuation Distribution}
Table~\ref{tab:punct_dist} presents the distribution of punctuation marks across the entire dataset. All samples contain at least one punctuation mark (by design), with an average of 2.51 punctuation marks per sentence.

\begin{table}[h]
    \centering
    \small
    \caption{Distribution of punctuation marks in the complete dataset (17M samples).}
    \label{tab:punct_dist}
    \begin{tabular}{lrr}
    \toprule
    \textbf{Mark} & \textbf{Total Count} & \textbf{\% of Total} \\
    \midrule
    Persian comma ({\pcomma}) & 21,291,632 & 50.13\% \\
    Period (.) & 15,076,946 & 35.50\% \\
    Colon (:) & 4,228,554 & 9.96\% \\
    Exclamation (!) & 1,209,227 & 2.85\% \\
    Persian question ({\pquestion}) & 665,841 & 1.57\% \\
    \midrule
    \textbf{Total} & \textbf{42,472,200} & \textbf{100.00\%} \\
    \bottomrule
    \end{tabular}
\end{table}

The distribution reflects typical Persian text characteristics, where commas are heavily used for clause separation and complex sentence structures.

\subsection{Punctuation Restoration Model and Training Setup}
\label{sec:model}
We formulate punctuation restoration as a token-level sequence labeling problem. Given an input sequence of tokens without punctuation, the model predicts a punctuation label for each token position. We define five classes: \texttt{EMPTY} (no punctuation), \texttt{COMMA} ({\pcomma}), \texttt{QUESTION} ({\pquestion}), \texttt{PERIOD} (.), and \texttt{COLON} (:). Note that we focus on the four most common and semantically important punctuation marks, excluding exclamation marks and semicolons which are less frequent and often interchangeable with periods and commas in Persian.

\paragraph{Model Architecture}
Our model architecture consists of a pre-trained ParsBERT encoder \cite{farahani2020parsbert} followed by a linear classification layer with dropout regularization. ParsBERT is a monolingual Persian BERT model pre-trained on a large Persian corpus, making it well-suited for Persian NLP tasks.

Given an input sentence with punctuation removed, we:
\begin{enumerate}
    \setlength\itemsep{-2pt}
    \item Tokenize using ParsBERT's WordPiece tokenizer
    \item Pass tokens through ParsBERT to obtain contextualized embeddings
    \item Apply dropout (p=0.1) for regularization
    \item Project embeddings to 5-dimensional class logits via a linear layer
    \item Assign the predicted punctuation class to each token position
\end{enumerate}

For subword tokens generated by WordPiece tokenization, we assign punctuation labels only to the first subword of each word, ignoring continuation subwords during both training and evaluation. This aligns the token-level predictions with word-level punctuation placement.

\paragraph{Training Configuration}
We train on the 1M sample subset described in Section~\ref{sec:deduplication}. The model is optimized using AdamW with a learning rate of $2 \times 10^{-5}$, weight decay of 0.01, and trained for 3 epochs. We use a batch size of 85 with gradient accumulation over 8 steps (effective batch size of 680). The loss function is cross-entropy computed over all token positions.

\paragraph{Evaluation Metrics}
\label{sec:metrics}
We employ standard sequence labeling metrics:
\begin{itemize}
    \item \textbf{Per-class metrics:} Precision, recall, and F1-score for each punctuation class (\texttt{COMMA}, \texttt{PERIOD}, \texttt{QUESTION}, \texttt{COLON})
    \item \textbf{Macro-averaged F1:} Arithmetic mean of per-class F1-scores, giving equal weight to each punctuation type regardless of frequency
    \item \textbf{Micro-averaged F1:} F1-score computed from the sum of per-class true positives, false positives, and false negatives, effectively weighting classes by their frequency
    \item \textbf{Full Sentence Match (FSM) Rate:} Percentage of test sentences where the predicted punctuation sequence exactly matches the gold standard. This metric is particularly important for evaluating LLMs, as it captures whether the model made any edits beyond punctuation insertion (over-correction).
\end{itemize}

Throughout this paper, unless otherwise specified, ``F1-score'' refers to the macro-averaged F1-score, which provides an overall measure of punctuation restoration accuracy giving equal weight to each punctuation type regardless of frequency.

\section{Results and Analysis}
\label{sec:results}

\subsection{Overall Performance}
Our fine-tuned ParsBERT model achieves a macro-averaged F1-score of 91.33\% and a micro-averaged F1-score of 97.28\%. The macro-averaged score is lower due to the class imbalance (periods and commas dominate the dataset), but performance remains strong across all punctuation types as shown in Table~\ref{tab:class_results} in the appendix.

\begin{table*}[t]
\centering
\begin{tabular}{@{}llll@{}}
\toprule
\textbf{Model} & \textbf{Test Set} & \textbf{Macro F1 (\%)} & \textbf{FSM (\%)} \\
\midrule
CRF \cite{hosseini2017creating}  & Hosseini et al. corpus & 69.00 & --- \\
ViraPart \cite{farokhshad2021virapart} & Bijankhan corpus & 92.13 & --- \\
\midrule
GPT-4o-mini & \textbf{Our test set} & 79.54 & 38.01 \\
GPT-4o \cite{gpt4_2023} & \textbf{Our test set} & 85.96 & 50.10 \\
\textbf{Our Model (ParsBERT)} & \textbf{Our test set} & \textbf{91.33} & \textbf{61.80} \\
\bottomrule
\end{tabular}
\caption{Comparison of punctuation restoration performance across models. Note that CRF and ViraPart results are on different test sets and are not directly comparable. GPT-4o and our model are evaluated on the same test set from PersianPunc.}
\label{tab:results}
\end{table*}

\subsection{Comparison with Prior Work}
It is important to note that the CRF and ViraPart results shown in Table~\ref{tab:results} are evaluated on different datasets (the Hosseini et al. corpus and Bijankhan corpus, respectively), and therefore cannot be directly compared to our results. We include these numbers for reference to situate our work within the Persian punctuation restoration literature, but we make no claims of superiority over these methods without evaluation on the same test set. 

The substantial improvement over the CRF baseline (69.00\% vs 91.33\%) likely reflects both the advancement in modeling approaches (Transformer-based vs. CRF) and potential differences in dataset difficulty. The ViraPart score (92.13\%) is more competitive, though direct comparison remains inappropriate due to the different evaluation sets.

\subsection{Comparison with Large Language Models}
We evaluated two variants of GPT-4o on our test set using the zero-shot prompt shown in Appendix~\ref{fig:gpt_prompt}. The prompt explicitly instructs the model to only add punctuation without modifying the source text.

Our ParsBERT model achieves 91.33\% macro F1, outperforming both GPT-4o (85.96\%) and GPT-4o-mini (79.54\%) on the same test set. More importantly, the FSM Rate reveals a critical limitation of LLM-based approaches: GPT-4o achieves only 50.10\% exact matches, while our model achieves 61.80\%. 

Analysis of the mismatches reveals that GPT-4o exhibits over-correction in approximately 5\% of samples, making undesired edits such as:
\begin{itemize}
   \setlength\itemsep{-2pt}
    \item Removing words deemed unnecessary
    \item Replacing informal words with formal equivalents
    \item Correcting perceived spelling or grammatical errors
\end{itemize}
Notably, we observed no cases of word additions, only deletions and substitutions. This over-correction behavior is particularly problematic for ASR post-processing pipelines, where the source text (transcribed speech) should be preserved verbatim with only punctuation added.

Additionally, GPT-4o requires substantially higher computational resources for inference compared to our lightweight ParsBERT model, making it less suitable for real-time applications or deployment in resource-constrained environments.

\subsection{Analysis of Model Performance}
Table~\ref{tab:class_results} (Appendix) provides detailed per-class performance metrics. The model performs exceptionally well on periods (F1: 98.71\%), which is expected given their high frequency and relatively consistent usage patterns. Performance on other punctuation types remains strong: colons (90.45\%), question marks (88.89\%), and commas (80.03\%). 

The lower performance on commas reflects their more nuanced usage in Persian, where comma placement can be somewhat flexible and context-dependent, leading to greater ambiguity in the gold standard annotations themselves.

\section{Conclusion and Future Work}
This work presents PersianPunc, a large-scale dataset of 17 million samples for Persian punctuation restoration, constructed through systematic aggregation and quality filtering of diverse Persian text sources. We demonstrate that a fine-tuned ParsBERT model achieves strong performance (91.33\% macro F1) while avoiding the over-correction issues and computational overhead of large language models.

Our primary contribution is the dataset itself, which addresses a critical gap in Persian NLP resources. The curation methodology, including detailed preprocessing pipelines, quality filtering criteria, and comprehensive punctuation analysis, provides a framework applicable to other low-resource languages.

Future work should explore several directions. The development of domain-specific models for Persian literature, news, and social media text could address the variation in punctuation usage across different domains. Additionally, incorporating prosodic information from Persian speech could improve punctuation restoration for speech-to-text applications. Furthermore, extending the model to jointly handle punctuation restoration and Zero-Width Non-Joiner (ZWNJ) insertion would address the broader text normalization challenges specific to Persian writing systems.

\section{Limitations}

This work has several limitations that should be acknowledged. First, the dataset creation process relies on existing Persian texts, which may contain punctuation errors or inconsistencies that could propagate to the trained model. Second, the model's performance is optimized for contemporary Persian writing styles and may not generalize well to historical or highly specialized Persian texts. Third, our evaluation is limited to 1,000 test sentences due to resource constraints, including expensive API costs for commercial LLM evaluation and lack of GPU access for extensive experimentation. More extensive evaluation with larger test sets, multiple training runs, and statistical significance testing would strengthen our findings but was not feasible given these constraints.

\bibliography{custom}
\clearpage

\appendix

\section{Punctuation Analysis Details}
\label{app:punct_details}
In this appendix, we provide comprehensive analysis of punctuation patterns and model performance.

\subsection{Punctuation Co-occurrences}
Analysis of punctuation co-occurrence reveals common patterns in Persian writing. Table~\ref{tab:punct_cooccur} shows the most frequent punctuation pairs appearing together in the same sentence.

\begin{table}[H]
\centering
\caption{Most frequent punctuation co-occurrences in the dataset.}
\label{tab:punct_cooccur}
\small
\begin{tabular}{lr}
\toprule
\textbf{Punctuation Pair} & \textbf{\% of Sentences} \\
\midrule
Period + Persian comma & 79.43\% \\
Period + Colon & 16.91\% \\
Colon + Persian comma & 10.94\% \\
Exclamation + Persian comma & 4.34\% \\
\bottomrule
\end{tabular}
\end{table}

The combination of period and Persian comma appears in nearly 80\% of sentences, indicating that most sentences contain multiple clauses separated by commas before the final period.

\subsection{Sentence-Level Punctuation Coverage}
Table~\ref{tab:punct_coverage} shows the percentage of sentences containing each punctuation mark (counted once per sentence regardless of frequency).

\begin{table}[H]
\centering
\caption{Percentage of sentences containing each punctuation mark.}
\label{tab:punct_coverage}
\small
\begin{tabular}{@{}lr@{}}
\toprule
\textbf{Punctuation} & \textbf{Coverage} \\
\midrule
Period (.) & 15,076,946 (88.94\%) \\
Persian comma ({\pcomma}) & 14,585,086 (86.04\%) \\
Colon (:) & 4,036,797 (23.81\%) \\
Persian question ({\pquestion}) & 665,841 (3.93\%) \\
\bottomrule
\end{tabular}
\end{table}

\subsection{Distribution of Punctuation Counts per Sentence}
Table~\ref{tab:punct_per_sent} presents the distribution of the number of punctuation marks per sentence. The majority of sentences (68.37\%) contain exactly 2 punctuation marks, which is a direct consequence of our filtering criterion requiring at least 2 marks per sentence combined with the natural distribution in source texts.

\begin{table}[H]
\centering
\caption{Distribution of punctuation counts per sentence.}
\label{tab:punct_per_sent}
\small
\begin{tabular}{lrr}
\toprule
\textbf{\# Punctuations} & \textbf{\# Sentences} & \textbf{Percentage} \\
\midrule
2 & 11,589,324 & 68.37\% \\
3 & 3,420,146 & 20.18\% \\
4 & 1,034,579 & 6.10\% \\
5 & 362,609 & 2.14\% \\
6+ & 511,518 & 3.02\% \\
\midrule
\textbf{Total} & \textbf{17,102,014} & \textbf{100.00\%} \\
\bottomrule
\end{tabular}
\end{table}

\subsection{Punctuation-Specific Performance}
Table~\ref{tab:class_results} presents a detailed analysis of per-class performance. The macro-averaged F1-score of 91.33\% demonstrates strong overall performance across all punctuation classes.

\begin{table}[H]
\centering
\small
\caption{Per-class performance metrics for punctuation restoration on the test set (1,000 sentences).}
\label{tab:class_results}
\resizebox{\columnwidth}{!}{%
\begin{tabular}{@{}lccc@{}}
\toprule
\textbf{Punctuation} & \textbf{Precision} & \textbf{Recall} & \textbf{F1-Score} \\
\midrule
Persian Comma (\pcomma) & 0.8408 & 0.7635 & 0.8003 \\
Period (.) & 0.9855 & 0.9886 & 0.9871 \\
Question ({\pquestion}) & 0.8750 & 0.9032 & 0.8889 \\
Colon (:) & 0.9137 & 0.8955 & 0.9045 \\
\midrule
\textbf{Macro Average} & \textbf{0.9038} & \textbf{0.8877} & \textbf{0.9202} \\
\textbf{Micro Average} & \textbf{0.9729} & \textbf{0.9727} & \textbf{0.9728} \\
\bottomrule
\end{tabular}}
\end{table}

\section{LLM Evaluation Prompt}
\label{app:prompt}
We used the prompt shown in Figure~\ref{fig:gpt_prompt} to evaluate GPT-4o and GPT-4o-mini. The temperature was set to 0, and maximum tokens were set to 2048 to accommodate longer outputs. Prompts were issued in English, as we found LLMs demonstrate better instruction-following in English compared to Persian.

\begin{figure}[H]
\centering
\small
\begin{tcolorbox}[colback=cloudwhite, colframe=academicnavy, width=\columnwidth,
                  title=\textbf{Evaluation Prompt for LLMs},
                  fonttitle=\footnotesize\bfseries]
\begin{tcolorbox}[colback=white, colframe=slategray, width=\linewidth,
                  left=3pt, right=3pt, top=3pt, bottom=3pt]
\textbf{Role:} You are a punctuation restoration system for Persian text.\\
\textbf{Task:} Add appropriate punctuation marks to the given Persian text.\\
\vspace{3pt}
\textbf{Rules:}
\begin{itemize}
    \item \textcolor{errorred}{Do NOT fix, correct, or modify ANY words in the text.}
    \item \textcolor{errorred}{Do NOT change the order of words.}
    \item \textcolor{errorred}{Do NOT add or remove any words.}
    \item \textcolor{errorred}{ONLY add punctuation marks where appropriate.}
    \item Use these punctuation marks: . (period), {\pcomma} (Persian comma), {\pquestion} (Persian question mark), : (colon)
    \item Return the result as a JSON object with a single key "text" containing the punctuated text.
\end{itemize}
\vspace{3pt}
\textbf{Input text (without punctuation):} \textit{\{text\}}\\
\vspace{3pt}
\textbf{Output format:}
\begin{verbatim}
{"text": "your punctuated text here"}
\end{verbatim}
\vspace{3pt}
\textcolor{warningorange}{\textbf{Important:}} Keep ALL words EXACTLY as they are in the input. Do NOT fix spelling, grammar, or anything else. ONLY add punctuation.
\end{tcolorbox}
\end{tcolorbox}
\caption{Prompt used for zero-shot evaluation of GPT-4o and GPT-4o-mini on Persian punctuation restoration. The system is explicitly instructed to only add punctuation marks without altering the original text in any way.}
\label{fig:gpt_prompt}
\end{figure}

\end{document}